%% file: pair2struct.tex
\documentclass{article} 
\usepackage{pair2struct,times}

\input{math_commands.tex}

\usepackage{hyperref}
\usepackage{url}

\usepackage{xspace}
\usepackage{caption}
\usepackage{subcaption}
\usepackage{amsmath}
\usepackage{multirow}
\usepackage[utf8]{inputenc} 
\usepackage[T1]{fontenc}    
\usepackage{booktabs}       
\usepackage{amsfonts}       

\usepackage{microtype}      
\usepackage{xcolor}         
\usepackage{graphicx}

\title{ConceptDistil: Model-Agnostic Distillation of Concept Explanations}

\author{João Bento, Ricardo Moreira, Vladimir Balayan, Pedro Saleiro, Pedro Bizarro\\
Feedzai\\
\texttt{\{firstname.lastname\}@feedzai.com} \\
}

\iclrfinalcopy 
\begin{document}

\maketitle
\begin{abstract}
Concept-based explanations aims to fill the model interpretability gap for non-technical humans-in-the-loop.
Previous work has focused on providing concepts for specific models (e.g, neural networks) or data types (e.g., images), and by either trying to extract concepts from an already trained network or training self-explainable models through multi-task learning.
In this work, we propose ConceptDistil, a method to bring concept explanations to any black-box classifier using knowledge distillation. ConceptDistil is decomposed into two components: (1) a concept model that predicts which domain concepts are present in a given instance, and (2) a distillation model that tries to mimic the predictions of a black-box model using the concept model predictions.
We validate ConceptDistil in a real world use-case, showing that it is able to optimize both tasks, bringing concept-explainability to any black-box model.
\end{abstract}

\section{Introduction}
\label{sec:introduction}

\begin{figure}[h]
    \begin{center}
        \includegraphics[width=0.95\textwidth]{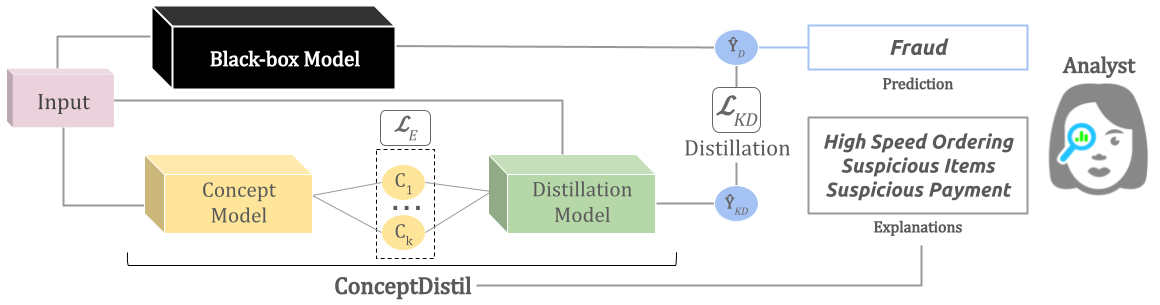}
    \end{center}
    \caption{ConceptDistil explaining a black-box applied to the fraud detection domain.}
    \label{fig:teaser}
\end{figure}

Mainstream methods to interpret black-box models produce explanations based on feature-attribution coefficients \citep{Lundberg2017, Ribeiro2016, Sundararajan2017}. While useful for a technical persona (\textit{e.g.}, a data scientist), these low-level explanations are difficult to grasp for most humans-in-the-loop (\textit{e.g.}, medical doctors, fraud analysts). %
%
Recently, concept-based explainability has been a promising line of research in which explanations are based on the importance of high-level semantic concepts instead of model features.  

Previous work on concept-based explainability, such as TCAV or ACE \citep{Kim2018, Ghorbani2019}, rely on \textit{post-hoc} learning of concepts and subsequent assessment of their importance for the prediction of a given class. Although relevant for a global understanding of a model, this family of methods does not help humans-in-the-loop making individual decisions (\textit{e.g.}, fraud analyst that discern an individual transaction). 


Another family of concept-based explainability methods aims to produce instance-level concept explanations by changing the model learning itself. These approaches use multi-task learning consisting of a main classification task and an explainability task~\citep{balayan2020teaching, Koh2020conceptbottleneck, li2018deep, melis2018towards, nanda2020unifying}. Although promising, these approaches attain sub-par performance in the main classification task due to known trade-offs in multi-task learning~\citep{sener2018multi}. 

The aforementioned approaches are model-specific as they they assume a structure based on neural networks (NN). However, for applications that work with tabular data, gradient boosting models are still quite popular ~\citep{borisov2021deep, shwartz2022tabular, shavitt2018regularization, gorishniy2021revisiting, popov2019neural}. This urges the need for a concept explainability method for any type of model.




In this work we propose ConceptDistil, illustrated in Figure~\ref{fig:teaser}, which introduces concept explainability as an independent component without changes to the classification model. It builds a distillation model that approximates the classifier while also producing concept explanations. The advantages of this approach are two fold: the classification model maintains its original performance, and concept explainability becomes model-agnostic, which is crucial in the context of tabular data. 

ConceptDistil is divided in two sequential components, a concept model which predicts which domain concepts are present on a given instance, and an attention-based distillation model that approximates the predictions of the black-box classifier. The distillation model explains the classifier as a weighted sum of the concept model's predictions. This offers two separate explanations types: (1) a data explanation that lists which concepts are present on a given instance and (2) a classifier explanation which gives the contributions of each concept to the classification score. 

We validate our method in a real-world fraud detection tabular dataset. We show that ConceptDistil is able to approximate the behaviour of the explained model while still achieving high performance on the explainability task. 

The main contributions of our work are:
\begin{itemize}
    \item We propose a method for concept explainability that does not compromise classification performances;
    \item Our method is model-agnostic which makes it applicable to a wider range of domains;
    \item Our method provides both data explanations and concept score contributions which correspond to model explanations.
\end{itemize}

\section{Concept-based knowledge distillation explainer}
\label{sec:surrogate_exp}
\textbf{Preliminaries.} We frame $D$ as a binary classification task, $\displaystyle \vy_D \in \sY_D = \{0, 1\}$ and $\mathcal{C}$ as a black-box binary classification model of the form $\displaystyle \mathcal{C}: \sX \rightarrow \sY_D$ where $\mathcal{C}(\vx) = \hat{\vy}_D$, where the random variable $\displaystyle (\vx, \vy_D)$ has an unknown joint distribution in $\sX \times \sY_D$ and $\hat{\vy}_D$ is the classifier estimate of $\vy_D$.
We want to explain $\mathcal{C}$ by distilling its knowledge into an explainer $f$ that jointly predicts domain concepts to serve as data explanations as well as the behaviour (score) of the explained model.

We frame our explainer as a multi-task surrogate model with two individual tasks: a knowledge distillation task, $KD$, where $f$ learns to approximate the classification scores of the black-box $\mathcal{C}$ given by $\displaystyle \vy_{KD} \in \sY_{KD} = [0, 1]$, and a multi-label classification task, $E$, with targets $\displaystyle \vy_E \in \sY_E = \{0, 1\}^K$, where $f$ learns to predict $k$ concepts which serve as the data explanations.
We define our explainer as $\displaystyle f: \sX \rightarrow \sY_{E} \times \sY_{KD}$, where $f(\vx) = (\hat{\vy}_E , \hat{\vy}_{KD})$ is the $(k+1)$-dimensional output vector of $k$ concept predictions plus the predicted knowledge distillation score.
We seek to maximize the explainer's performance in both tasks during training.
Let $\displaystyle \Ls_{KD}(\hat{\vy}_{KD}, \mathcal{C}(\vx))$ and $\displaystyle \Ls_E(\hat{\vy}_E, \vy_E)$ represent the losses incurred by the model at the knowledge distillation and explainability tasks, respectively.
The surrogate explainer model $f$ minimizes the weighted combination of both losses, as defined by:
\begin{equation}
\label{eq:final_loss}
\displaystyle \Ls(\hat{\vy}, \vy) = \lambda \, \Ls_{KD}(\hat{\vy}_{KD}, \mathcal{C}(\vx)) + (1 - \lambda) \, \Ls_E(\hat{\vy}_E, \vy_E)
\end{equation}
Given the inherent fidelity-explainability trade-off, we may assume $\displaystyle \lambda \in [0, 1]$ as a hyperparameter to weight the relative importance of the knowledge distillation task with respect to the explainability task. A high $\lambda$ allows for a higher fidelity of $f$ to the actual predictions of $\mathcal{C}$ at the cost of lower performance on the explainability task. On the contrary, a low $\lambda$ allows for $f$ to have an higher explainability performance but at the risk of it being less coupled with the black-box model $\mathcal{C}$. We show the effect of this parameter on appendix \ref{appendix:hyperparams_lambda}.
%
Depending on the nature of each task, different loss functions may be used. In this work, we opted for the Binary Cross-Entropy (BCE) loss as the $\displaystyle \Ls_{KD}$ and used the mean BCE loss over the $K$ concepts as the $\displaystyle\Ls_E$. More formally: 
\begin{equation}
\displaystyle\Ls_E(\hat{\vy}_E, \vy_{E}) = \frac{1}{K}\sideset{}{_{i=0}^K}\sum \text{CE}(\hat{\vy}_E^{(i)}, \vy_E^{(i)})   
\end{equation}

\begin{figure}[h]
    \begin{center}
        \includegraphics[width=0.85\textwidth]{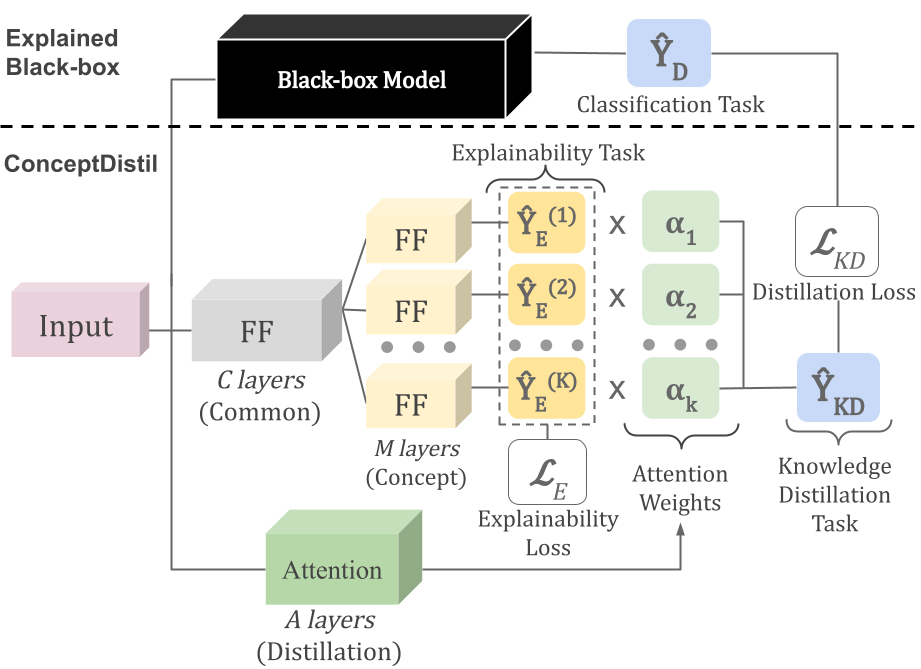}
    \end{center}
    \caption{Diagram of ConceptDistil architecture and its main components.}
    \label{fig:concepttab}
\end{figure}

We propose the explainer architecture illustrated in Figure~\ref{fig:concepttab}, composed by $C$ feedforward (FF) layers, parameterized by $\vtheta_{C}$, which are shared across all concepts, $M \times K$ layers, parameterized by $\{\vtheta_{M}^{(i)}\}_{i=1}^{K}$ which are concept-specific, and $A$ layers, parameterized by $\vtheta_{A}$, which are responsible for the attention-based distillation task.

The output of the explainability task is $\displaystyle \hat{\vy}_E$, which can be divided into $K$ components for each concept $i$, of the form: 
\begin{equation}
\displaystyle \hat{\vy}_E^{(i)} = h(h(\vx; \vtheta_{C}); \vtheta_{M}^{(i)})
\end{equation}
Where $\displaystyle h(\vx; \vtheta)$ represents the neurons activation resulting from inputting $\displaystyle \vx$ into a network parameterized by $\displaystyle \vtheta$.
%
The output of the knowledge distillation task is $\hat{\vy}_{KD} = \sum_{i=0}^{K} \hat{\vy}_E^{(i)} \times \alpha_i$, where $\alpha_i$ represents the attention coefficient for the $i$-th concept, calculated by a simple attention mechanism~\citep{bahdanau2014neural}. This attention mechanism is parameterized by $\vtheta_{A}$ and receives as input the same input $\vx$ as the black-box model and the concept model. The attention coefficients are then defined as:

\begin{gather}
    \alpha_i = softmax(e_i, e)\\
    e = h(\vx; \vtheta_A)
\end{gather}

\textbf{Model and Data Explainability.} The aforementioned architecture, illustrated on Figure~\ref{fig:concepttab}, is therefore a composition of two models: (1) a concept model, consisting of $C$ common layers and $M \times K$ concept-specific layers, that predicts which concepts are present on a given input; and (2) a distillation model, composed of $A$ layers, which, for a given input, attributes attention-based importances to the predicted concepts in order to obtain the black-box assigned score.
Considering the two models, our method produces two types of explanations which can be provided to users:
(1) the concepts present on a given input, as predicted by the concept model;
(2) how much each concept prediction is relevant to obtain the final black-box model output score.

\textbf{Task Contamination.} One problem that arise from jointly learning \textbf{ConceptDistil} is the problem of task contamination. This phenomena happens when, to improve the knowledge distillation task, the gradients force the already correctly learned concept model to degrade its explainability performance. 
To erase the possibility of task contamination we propose an alternative, named \textbf{ConceptDistil -- No gradient}, where the propagation of gradients back from the attention mechanism into the concept model is blocked. This allows ConceptDistil to separate the knowledge distillation and the explainability tasks while training.
We further propose another alternative to overcome task contamination, dubbed as \textbf{ConceptDistil -- 2-staged}, where we train the two components of ConceptDistil in separate stages. In this training framework, the concept model is trained on the concept labels, and only after is the distillation model trained using as inputs the predictions of the already trained concept model.

\section{Experimental setup}
\label{sec:experimental_setup}

We validate our method in real world financial fraud detection application where the data is typically in the tabular form. In our use-case, predictive performance is often the single metric of interest. Therefore, it is crucial to deploy the model with maximum performance.
Thus, we ran an experiment (details in Appendix \ref{appendix:tabular_performance}) to find models with higher performance for the predictive task. Therefore, we decided to evaluate our ConceptDistil approaches using the best LightGBM and FeedForward Neural Network (FFNN) models as the base classifiers.


\textbf{Dataset.} For our experiments, we used a privately held online retailer fraud detection dataset, comprised of approximately 5.5M transactions where 2\% represent fraudulent behaviour. The dataset is then divided into three sequential subsets: train, validation and test, composed of 4.8M, 200k and 1M instances respectively. Because our goal is model explainability, we created a sample of the transactions which would most likely need to be explained, containing the transactions of highest uncertainty for the historical fraud detection system. This sample resulted in 460k instances for training, 17k for validation and 107k for test. Our ConceptDistil models were trained and tested using these sampled datasets.


\textbf{Golden Concept Labels.} To evaluate our explainability task, we randomly sampled a small set from the sampled dataset, and presented those transactions to a group of human experts (fraud analysts with the knowledge of the most common fraud patterns latent on this online retailer dataset). We then asked the fraud analysts to annotate the existing concepts on each transaction. We present more details about this golden concept dataset in Appendix Section~\ref{appendix:golden_dataset}


\textbf{Concept Weak Supervision.} To train our explainability models using the full and sampled datasets we employed a weak supervision methodology. This consisted of creating a set of \textit{Concept Teachers} of the form $\displaystyle f: \sX \rightarrow \sY_{E}$ trained using a small golden training set with concept labels obtained with the method described above. The \textit{Concept Teachers} were then used to infer the probabilistic labels $p(\vy_E^{(i)}|\vx)$ of the concept $i$ being present in transaction $\vx$. Our explainability models were trained using these probabilistic labels as targets. We further detail how we developed our \textit{Concept Teachers} in the Appendix Section~\ref{appendix:exp_setup}.

\textbf{Hyperparameter Optimization.} We selected the best hyperparameters using a Tree-structured Parzen Estimator (TPE) \citep{bergstra2011algorithms} optimization of 50 trials for each variant. We further detail the hyperparameters in appendix \ref{appendix:hyperparams}.


\textbf{Evaluation Metrics.} To evaluate the knowledge distillation task we used a measure of fidelity derived from the Mean Absolute Error (MAE) given by:
\begin{equation}
fidelity = 1 - MAE = 1 - \frac{\sum_{i=1}^{n} |\hat{y}_{KD}^{(i)}-y_{KD}^{(i)}|}{n}, 
\end{equation}
where $n$ is the total number of instances in the evaluation set. Because our knowledge distillation task approximates a classification score between 0 and 1, we can also see that this measure is always within the same interval.
To evaluate the explainability task, because we do not wish to focus on any specific range of False Positive Rate (FPR), we measure the area under the ROC curve (ROC AUC) of each individual concept. We then take the average of these measures across concepts to obtain a single evaluation measure for the explainability task. The explainability task evaluation is performed on the small golden set of human-annotated concept labels.

\section{Results}

In total, we trained around 1600 models, and summarized the obtained results in the Table \ref{tab:results_table}. Along with our ConceptDistil models, we created 2 baseline models which help us understand the performance bounds on the two individual tasks. We detail an implementation for those baselines in appendix \ref{appendix:baselines}.



The results of Table \ref{tab:results_table} show a clear difference in distillation performance between the two black-box classifiers used. The fidelity performance achieved when distilling the FFNN model is higher for all surrogate architectures tested. We believe that the distillation of the FFNN is a simpler task when compared with the distillation of LightGBM due to the similarity of inductive biases between the FFNN classifier and the surrogate architectures. 

\textbf{Baselines comparison.} We can see in Table~\ref{tab:results_table} that our best Distillation baseline achieved 93.31\% fidelity when distilling the LightGBM classifier and 98.44\% when distilling the FFNN classifier, which is superior to the best fidelity obtained by ConceptDistil explainers (91.43\% for LightGBM and 97.10\% for FFNN), showing approximately a 1.9 and 1.3 percentage points (pp) gap in fidelity for the LightGBM and FFNN distillation, respectively. The biggest gap in fidelity (approximately 6.1 pp) occurred for the ConceptDistil -- \textit{2-staged} when distilling the LightGBM classifier. In terms of the explainability baseline, the best model attains 78.12\% of mean ROC AUC. Since the ConceptDistil -- \textit{2-staged} uses this explainability baseline as its own data explainer, its explainability performance is the same as the baseline. In contrast, for jointly learned ConceptDistil models, there is a drop of 1.6 pp and 1 pp in explainability performance when distilling LightGBM and FFNN, respectively.
These results confirm the existence of a trade-off between knowledge distillation and explainability tasks since both single-task baselines outperform the proposed multi-task approaches.

\begin{table}[htbp]
\centering
 \caption{Performance comparison between different concept-based knowledge distillation explainers.}
\begin{tabular}[t]{ll|cc|cc}
\toprule
\multicolumn{2}{c}{} & \multicolumn{2}{c}{LightGBM} & \multicolumn{2}{c}{FFNN} \\
\midrule
& & Fidelity & Mean & Fidelity & Mean\\
& & (\%) & ROC AUC (\%) & (\%) & ROC AUC (\%)\\
\midrule
\multirow{2}{5em}{Baselines} & Distillation Model & \textbf{93.31} & - & \textbf{98.44} & - \\
& Explainability Model  & - & \textbf{78.12} & - & \textbf{78.12}\\
\midrule
\multirow{3}{5em}{ConceptDistill}& Default & \textbf{91.43} & 76.43 & \textbf{97.10} & 77.16\\
& 2-staged & 88.07 & \textbf{78.12} & 94.73 & \textbf{78.12} \\
& No gradient & 87.20 & 76.28 & 95.26 & 77.58\\
\bottomrule
 \end{tabular}
\label{tab:results_table}
\end{table}

\textbf{ConceptDistil -- 2-staged vs ConceptDistil -- No gradient.} When comparing the results of these variants, we see that, when distilling the LightGBM, the \textit{No gradient} variant performs worse than the \textit{2-staged} on both tasks. When distilling the FFNN, the \textit{No gradient} achieves better distillation but worse explainability which corresponds to a softer trade-of between distillation and explainability than the \textit{2-staged}.

\textbf{ConceptDistil -- 2-staged/No gradient vs ConceptDistil.} When comparing \textit{2-staged} and \textit{No gradient} variants with the vanilla ConceptDistil, we see that the \textit{2-staged} achieves higher explainability performance while vanilla ConceptDistil attains better fidelity when distilling the LightGBM model. For the FFNN, vanilla ConceptDistil attains better fidelity, without much sacrifice in explainability performance (0.96 pp loss).
%

\section{Conclusion}

We present ConceptDistil, a concept-based knowledge distillation method that provides both data and model explanations. ConceptDistil uses a surrogate neural network that approximates the predictions of a black-box classifier while simultaneously producing concept explanations. As our method operates independently from the black-box model, it is model-agnostic and does not compromise the classification task performance. We validated ConceptDistil on a privately held real-world online retailer fraud detection dataset, and showed that ConceptDistil can learn both the explainability and the distillation tasks.

%
%
%



\bibliography{pair2struct}
\bibliographystyle{pair2struct}

\appendix
\section{Appendix}

\subsection{Predictive performance on real world financial crime dataset}
\label{appendix:tabular_performance}

We ran experiments using private real-world online retailer fraud detection dataset that contains approximately 5.5M transactions. This dataset is highly unbalanced, with only 2\% of transactions being fraudulent. We used TPE as optimization algorithm, running 50 iterations for each model type. To evaluate the performance of models on the classification task (fraud task) we consider the True Positive Rate (TPR) (or \textit{Recall}), at 5\% FPR. This percentage of FPR (5\%) were picked due the business constrains defined by online retailer, minimizing the probability to harm an legitimate customer. The fraud performance evaluations were done using the full test dataset (not sampled) since we don't want to focus solely on the subset of transactions which require good explanations (FP, FN, uncertain) but instead on the whole set of transaction upon which the fraud model would take an approve/decline/review decision. The results for the best model of each type are presented in Table~\ref{tab:tabular_results}.


\begin{table}[htbp]
\centering
 \caption{Performance comparison between gradient-boosting-based, FFNN, and self-explainable models (\citep{balayan2020teaching, Koh2020conceptbottleneck}).}
\begin{tabular}[t]{lccc}
\toprule
\textbf{Model} & \textbf{Recall at 5\% FPR (\%)}  & \textbf{Out-of-the-box explanations}\\
\midrule
LightGBM  & 68.31 & Feature Importance Attributions &\\
FFNN & 65.55 & - \\
Self-explainable & 60.59 & Concepts (through Concept bottleneck) \\
\bottomrule
 \end{tabular}
\label{tab:tabular_results}
\end{table}

As already mentioned in section \ref{sec:introduction}, there are several studies which show that Deep Learning models struggle to outperform gradient-boosting-based models in tabular data. In this specific fraud detection use-case, we corroborate this finding with LightGBM model~\citep{ke2017lightgbm} achieving 68.31\% recall at the 5\% FPR level, while FFNN achieved 65.55\% recall at 5\% FPR. Additionally, we tested a concept self-explainable model~\citep{balayan2020teaching, Koh2020conceptbottleneck}, which achieved worse performance than both the best LighGBM and the FFNN, achieving 60.59\% recall at 5\% FPR. In this work, we test our ConceptDistil approaches using the best LightGBM and FFNN models as the base classifiers.

\subsection{Concept Golden Dataset}
\label{appendix:golden_dataset}
In section~\ref{sec:experimental_setup} we briefly describe the golden dataset of concepts that we use in our experiments. This dataset was obtained by collecting concept annotations in a total of 2643 transactions. These were further split into 1934 for training the \textit{Concept Teachers}, 203 for the optimization of the teachers' parameters using TPE, and 506 for test. This last set was the one used to evaluate the explainability performance of the \textit{Concept Teachers} and all ConceptDistil models and baselines presented. Despite the requirement for having concept labels being a limitation in this work, we can see that the effort of concept labelling can be mitigated by resourcing to a weak supervision strategy as the one used. We reduce the labeling effort in more than 3 orders of magnitude when compared to the effort of fully labeling the sampled dataset of around 500k transactions. Additionally, and despite the effort drawbacks, we consider that concept explainability can benefit from being establishing with ground truth concepts obtained from expert knowledge instead of unsupervised methodologies which can more easily result in meaningless concept definitions.

\subsection{Concept Teachers}
\label{appendix:exp_setup}

As described in section~\ref{sec:experimental_setup} our Weakly Supervised Concept Labeling consisted in learning a set of \textit{Concept Teachers}, which generated concept labels for all the remaining unlabeled transactions in the full and sampled datasets. To train the \textit{Concept Teachers} we used the golden dataset enriched not only with the fraud features available at run-time but also with some additional information, such as fraud rule triggers and manual decisions, which is only generated during or after the historical fraud decision. The algorithm used for the \textit{Concept Teachers} was a Random Forest classifier and the hyperparameters were optimized using a Tree-structured Parzen Estimator (TPE) \citep{bergstra2011algorithms} algorithm with 200 trials from which 30 trails were used for a random initialization. As the optimization metric we used ROC AUC on the golden validation set. We then evaluated the \textit{Concept Teachers} in the test set of our golden dataset obtaining a mean ROC AUC of 75.72\% across all concepts. Despite not being central to our work, the \textit{Concept Teachers} performance shows and interesting result when compared to the 78.12\% performance of our concept explainabilty baseline. This 2.4 pp improvement originates solely from the using more data in $\mathbb{X}$ (the concept explainability baselin trained in 460k instances labeled by our \textit{Concept Teachers}) which illustrates that our weak supervision approach is successful to some degree. 

\subsection{Concept Prevalences}
\label{appendix:concept_prevs}

To provide more context about the concepts considered in this work for the Fraud Detection, we show in Table \ref{tab:concept_prevs_table} the list of concepts as well as their prevalences on the golden dataset (Concept Teachers training data plus the golden test set).

\begin{table}[htbp]
\centering
 \caption{Concepts for the Fraud Detection use-case and corresponding prevalences on the golden dataset. The table shows the global prevalence and the prevalences conditioned on fraud classification labels (legit and fraud).}
\begin{tabular}[t]{lccc}
\toprule
\textbf{Concept} & \textbf{Global (\%)}  & \textbf{Legit (\%)} & \textbf{Fraud (\%)} \\
\midrule
Good Customer History  & 24.45 & 27.85 & 14.29 \\
High Speed Orderning & 11.33 & 8.49 & 19.84 \\
Suspicious Delivery & 22.86 & 20.16 & 30.95 \\
Suspicious Device & 11.73 & 8.75 & 20.63 \\
Suspicious Email & 21.07 & 18.93 & 30.16 \\
Suspicious Items & 18.49 & 17.24 & 22.22 \\
\bottomrule
 \end{tabular}
\label{tab:concept_prevs_table}
\end{table}

\subsection{Hyperparameter search}
\label{appendix:hyperparams}

For the hyperparameter optimization, we varied the number of $C$ layers between 3 and 5, the number of $M$ layers between $[3,7]$ and the number of $A$ layers between 1 and 4 as well as their respective layer dimensions between 2 and 2048. We let our loss trade-off parameter $\lambda$ vary between 0.2 and 0.8, and the learning rate vary between 0.0005 and 0.01. As regularization, we let all hidden layers dropout probabilities vary between 0 and 0.4, the l2 weight penalization parameter vary between 0 and 0.1 and let the optimizer select whether to use or not Batch Normalization \citep{ioffe2015batch}.

\subsubsection{Effect of $\lambda$}
\label{appendix:hyperparams_lambda}

\begin{figure}[h]
    \begin{center}
        \includegraphics[width=0.7\textwidth]{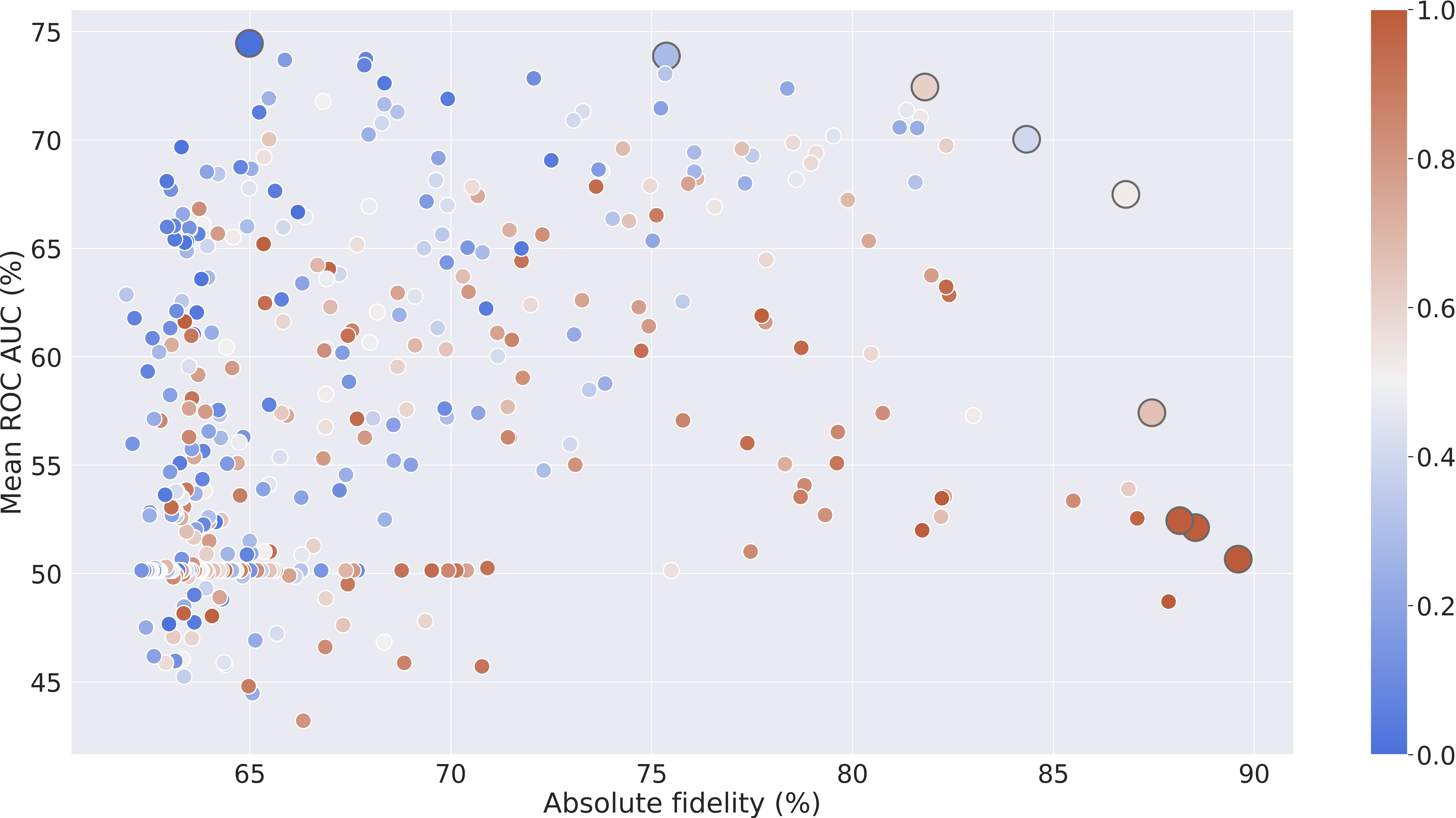}
    \end{center}
    \caption{The effect of $\lambda$ parameter on  Absolute fidelity (Knowledge Distillation task) and Mean ROC AUC (Explainability task). Bigger size points are on the Pareto frontier \cite{pareto1906manuale}.}
    \label{fig:lambda_effect}
\end{figure}

To understand the effect of the $\lambda$ parameter used in the loss function (equation \ref{eq:final_loss}), we ran an experiment where we test jointly learning ConceptDistil on best LightGBM model (\ref{appendix:tabular_performance}) with same hyperparameter space described in previous paragraph, with small difference for $\lambda$ that now vary between $[0,1]$. Instead of using TPE for hyperparameter optimization which may be biased due to its sequential optimization approach, we opt for random search. In total we trained around 400  models, and present the result on figure \ref{fig:lambda_effect}.

By analysing the bottom left area of the figure \ref{fig:lambda_effect} (absolute fidelity higher than 85\%), we observe that higher values of $\lambda$, close to 1, lead to higher performance on Knowledge Distillation task. On the contrary, the top left area (Mean ROC AUC higher than 70\%) is mainly populated with models that have small lambda values (close to 0). Thus, these models have higher weight on Explainability task, leading to better Mean ROC AUC values, but sacrificing absolute fidelity. The models that have a $\lambda$, values closed to 0.5, seems to achieve the better trade-off between two tasks, and are dominant on the Pareto frontier.

\subsection{Baselines}
\label{appendix:baselines}

We develop two baseline models that are specialized on two individual tasks (explainability and distillation).

\textbf{Concept Explainabilty Baseline.} Our first baseline is a model with the architecture of the concept model from ConceptDistil which, given an input $\vx$, predicts the probability $p(\vy_E|\vx)$. When evaluated on the golden test set, this model gives us an upper bound of the explainability performance. Additionally, the best model obtained for the concept model baseline is the one used for the ConceptDistil -- 2-staged. 

\textbf{Knowledge Distillation baseline} The second baseline model is a standard surrogate (dubbed as Distillation baseline) which, given an input $\vx$, predicts the output of the black-box classifier $\vy_{KD}$. This standard surrogate has a multi-layer FFNN architecture where the number of layers were varied between 4 and 9 and the layer dimensions were varied between 2 and 4096. When evaluated on the sampled test set, this standard surrogate model gives us an upper bound for the knowledge distillation task performance.

The hyperparameters of both baseline models were optimized using the same methodology and ranges described in appendix \ref{appendix:hyperparams}.

\end{document}

%% file: math_commands.tex
\usepackage{amsmath,amsfonts,bm}

\def\eqref#1{equation~\ref{#1}}

\def\1{\bm{1}}

\def\vtheta{{\bm{\theta}}}

\def\vx{{\bm{x}}}
\def\vy{{\bm{y}}}

\DeclareMathAlphabet{\mathsfit}{\encodingdefault}{\sfdefault}{m}{sl}
\SetMathAlphabet{\mathsfit}{bold}{\encodingdefault}{\sfdefault}{bx}{n}

\def\sX{{\mathbb{X}}}
\def\sY{{\mathbb{Y}}}

\newcommand{\Ls}{\mathcal{L}}

